%% file: main.tex
\documentclass[runningheads]{llncs}

\usepackage{eccv}

\usepackage{eccvabbrv}

\usepackage{graphicx}
\usepackage{booktabs}

\usepackage{multirow}
\usepackage{color}
\usepackage{colortbl}

\usepackage{algorithm}
\usepackage{algpseudocode}

\usepackage{amsfonts}
\usepackage{bbm}
\usepackage{bm}
\usepackage{wrapfig,lipsum}
\usepackage[utf8]{inputenc}
\usepackage{url}
\usepackage{booktabs}
\usepackage{amssymb}
\usepackage{bbding}
\usepackage{pifont}
\usepackage{wasysym}
\usepackage{utfsym}
\usepackage{amsmath}
\usepackage{amssymb}
\usepackage{fontawesome}
\usepackage{dashrule}

\usepackage{ragged2e}
\usepackage{graphicx}
\usepackage[table]{xcolor}
\usepackage[accsupp]{axessibility}  
\usepackage{xfp}
\usepackage{siunitx}
\newcommand{\avgthree}[3]{%
  \num{\fpeval{((#1)+(#2)+(#3))/3}}%
}

\definecolor{my_blue}{HTML}{d3eaf2}
\definecolor{Green}{rgb}{0.85882353, 0.90980392, 0.84705882}
\definecolor{rose}{rgb}{0.60392157, 0.53333333, 0.43921569}
\definecolor{dred}{rgb}{0.7254902, 0.09803922, 0.10588235}
\definecolor{VAIL_Green}{rgb}{0, .7, .0}

\newcommand{\pub}[1]{\color{gray}{\tiny{[{#1}]}}}

\definecolor{change_color}{RGB}{25,140,140}

\newcommand{\cmark}{\ding{51}} 
\newcommand{\xmark}{\textcolor{gray!40}{\ding{55}}}

\definecolor{citecolor}{rgb}{0.10,0.25,0.70}
\definecolor{urlcolor}{rgb}{0.00,0.30,0.70}
\usepackage[breaklinks,colorlinks,citecolor=citecolor,linkcolor=citecolor,urlcolor=urlcolor]{hyperref}
\usepackage{orcidlink}

\begin{document}

\title{CloSeR: Unified Relational Distillation from Closed-Set Teachers for Category Discovery} 

\titlerunning{CloSeR}


\author{
Yuanpei Liu\orcidlink{0009-0008-6144-6547} \and 
Zhenqi He\orcidlink{0009-0000-2265-7159} \and 
Jialu Tang\orcidlink{0009-0005-9861-9091} \and 
Kai Han\textsuperscript{\dag}\orcidlink{0000-0002-7995-9999}
}

\authorrunning{Y. Liu et al.}

\institute{Visual AI Lab, The University of Hong Kong, Hong Kong SAR \\
\email{ypliu0@connect.hku.hk, kaihanx@hku.hk}
}

\maketitle
\renewcommand{\thefootnote}{}
\footnotetext{\textsuperscript{\dag} Corresponding author.}

\input{secs/0_abstract}    
\input{secs/1_intro}
\input{secs/2_related}
\input{secs/3_method}
\input{secs/4_experiment}
\input{secs/5_conclusion}

%
%
\bibliographystyle{splncs04}
\bibliography{main}
\end{document}

%% file: secs/0_abstract.tex
\begin{abstract}
Generalized Category Discovery (GCD) is an intriguing open-world problem that has garnered increasing attention: given partially labelled data, the goal is to correctly recognize known classes while discovering coherent novel categories from unlabelled samples. Recent GCD methods typically adapt foundation models by jointly optimizing supervised classification and unsupervised discovery objectives on mixed labelled--unlabelled data. While effective, this coupled training can entangle closed-set recognition and open-set discovery, leading to objective conflict and biased predictions, and may disturb the semantic geometry of pretrained representations under limited labels and noisy pseudo-labels.
We propose \textbf{CloSeR}, a simple plug-and-play framework that injects \textbf{Clo}sed-\textbf{Se}t \textbf{R}elational knowledge into GCD training. CloSeR first builds a domain-adapted closed-set teacher by tuning lightweight block-wise adapters on labelled known-class data while keeping the foundation model backbone frozen, thereby preserving pretrained priors at low training cost. It then transfers the teacher’s knowledge to downstream GCD via {Unified Relational Distillation (URD)}, which distills complementary global sample-to-prototype relations to anchor known-class semantics and local sample-to-sample relations to preserve neighborhood structure, using separate feature pathways to reduce optimization interference. CloSeR is head-agnostic and readily integrates with both parametric and non-parametric GCD methods. Extensive experiments with DINO and DINOv2 backbones on six benchmarks (CIFAR-10/100, ImageNet-100, CUB, Stanford-Cars, and FGVC-Aircraft) show consistent gains over GCD baselines, achieving state-of-the-art performance.
Project page: \url{https://visual-ai.github.io/closer/}
\keywords{Generalized Category Discovery \and Self-Supervised Learning \and Transfer Learning
\and Knowledge Distillation}
\end{abstract}

%% file: secs/1_intro.tex
\section{Introduction}
\label{sec:intro}

\begin{figure}[t]
    \centering
    \includegraphics[width=\textwidth]{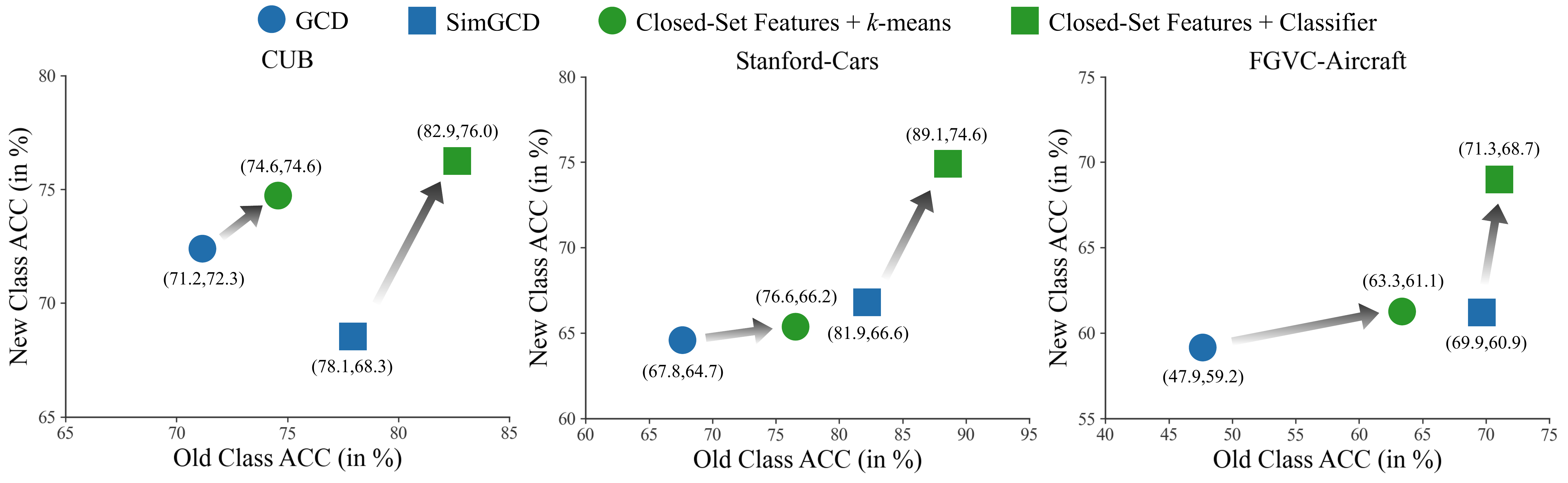}
    \caption{
    The \emph{Old}-class \emph{vs.} \emph{New}-class \textit{ACC} on CUB~\cite{wah2011caltech}, Stanford-Cars~\cite{krause20133d}, and FGVC-Aircraft~\cite{maji2013fine} using DINOv2~\cite{oquab2023dinov2} backbone. 
    We compare two representative GCD baselines (the \textit{non-parametric} GCD~\cite{vaze2022generalized} and the \textit{parametric} SimGCD~\cite{wen2023parametric}) with two simple alternatives that first adapt the backbone on labelled known classes (closed-set features) and then perform GCD via either $k$-means clustering or learning a parametric classifier. 
    \emph{Note:} ``GCD'' here denotes the method proposed in~\cite{vaze2022generalized}, not the task name.
    }
    \label{fig:intro}
\end{figure}

Once framed as Novel Category Discovery (NCD)~\cite{han2019learning} and later broadened to Generalized Category Discovery (GCD)~\cite{vaze2022generalized}, \textit{category discovery}~\cite{he2025category} has become a focal open-world challenge.
In GCD, we are given a partially-labelled dataset: the unlabelled split contains samples from both known and unknown categories.
The goal is to transfer knowledge from the labelled subset so that samples from known classes are correctly recognized while samples from unknown classes are grouped into their underlying novel categories.

Existing GCD methods can be broadly categorized by how they produce category predictions.
\textit{Non-parametric} approaches~\cite{vaze2022generalized,hao2023cipr,rastegar2023learn,RastegarECCV2024} discover categories by clustering features, whereas \textit{parametric} methods~\cite{wen2023parametric,wang2024sptnet,liu2025debgcd,he2025category} learn an explicit classifier (often prototype-based) to assign category indices.
With the rise of foundation models~\cite{caron2021emerging,oquab2023dinov2,radford2021learning}, recent works from both families typically adapt a pretrained backbone (via full fine-tuning or parameter-efficient tuning) and optimize supervised and unsupervised objectives jointly on the mixed labelled--unlabelled data.
While effective, this \emph{one-stage} training entangles known-class classification and novel-class discovery, which may introduce optimization conflict and prediction bias~\cite{liu2025debgcd,guo2025towards}.
At the same time, prior studies~\cite{vaze2022generalized,wang2024sptnet,liu2025hyperbolic} emphasize that GCD is essentially a \textit{transfer clustering} problem, where learning strong categorical priors from known classes is crucial.
Moreover, modern foundation models are pretrained on large-scale data and exhibit strong transferability and semantic structure~\cite{caron2021emerging,oquab2023dinov2}.
In principle, such pretrained priors should be highly beneficial for \emph{transfer clustering} in GCD; however, na\"ively adapting these models with mixed supervised--unsupervised objectives can distort the pretrained geometry, especially under limited labelled data and noisy pseudo-labels.
This suggests that \emph{how} we adapt foundation models may be as important as \emph{whether} we adapt them.
Taken together, these considerations raise a natural question:
\textit{Can we explicitly learn to recognize known classes first, and then leverage this closed-set knowledge to guide category discovery?}

To answer this, we conduct a simple pilot study.
We \emph{fully fine-tune} a foundation model (\eg, DINOv2~\cite{oquab2023dinov2}) on labelled known classes using a standard cross-entropy objective with a \emph{small learning rate} to preserve the pretrained prior and mitigate overfitting, and then evaluate the resulting representations using either vanilla $k$-means clustering or a lightweight parametric classifier trained on top of the frozen backbone following the standard objective~\cite{wen2023parametric}.
As shown in Fig.~\ref{fig:intro}, both \textit{non-parametric} and \textit{parametric} solutions improve markedly compared to their standard GCD counterparts.
This suggests that \emph{explicitly consolidating categorical priors from known classes} can significantly enhance the representation quality for category discovery.
However, this na\"ive two-step strategy is limited in practice:
\textbf{(i)} fully fine-tuning foundation models is expensive, especially for larger variants (\eg, ViT-L), which limits the scaling up of GCD models, and
\textbf{(ii)} simply freezing the adapted backbone and training a separate GCD head is inflexible and does not readily extend to diverse GCD pipelines, particularly non-parametric methods that rely on end-to-end feature learning.

We therefore propose \textbf{CloSeR}, a simple yet effective framework that injects \textbf{Clo}sed-\textbf{Se}t \textbf{R}elational knowledge into GCD training in a plug-and-play manner.
CloSeR consists of two stages.
\textbf{First}, we perform {Closed-Set Transfer Learning (CST)} by inserting \emph{lightweight block-wise adapters}~\cite{chen2022adaptformer} into a \emph{frozen} foundation model and learning closed-set prototypes using \emph{labelled data only}.
By updating only a small set of adapter parameters, CST provides an \emph{efficient} way to adapt shallow-to-deep features, yielding a domain-adapted closed-set teacher that preserves the strong pretrained prior while bridging the target-domain gap at low training cost.
\textbf{Second}, during GCD training, we introduce {Unified Relational Distillation (URD)}, which transfers \textbf{(i)} \textit{global} sample-to-prototype relations to anchor semantics to known-class prototypes and \textbf{(ii)} \textit{local} sample-to-sample relations to preserve neighborhood geometry.
To reduce optimization interference, URD decouples the feature pathways used for global and local relations.
Importantly, CloSeR can regularize both \textit{parametric} and \textit{non-parametric} GCD methods: it provides relational supervision to stabilize representation learning, regardless of whether category indices are produced by a classifier head or by clustering.

We evaluate CloSeR with different pretrained backbones (DINO and DINOv2) on six benchmarks, including CIFAR-10~\cite{krizhevsky2009learning}, CIFAR-100~\cite{krizhevsky2009learning}, ImageNet-100~\cite{deng2009imagenet}, CUB~\cite{wah2011caltech}, Stanford-Cars~\cite{krause20133d}, and FGVC-Aircraft~\cite{maji2013fine}.
Across GCD baselines spanning both \textit{parametric} and \textit{non-parametric} methods, CloSeR yields consistent improvements.
In summary, we make the following contributions:
\textbf{(i)} We identify a key limitation of the \textit{de facto} GCD training pipeline and show that {adapting foundation models to known classes before discovery} is crucial for robust category discovery.
\textbf{(ii)} We propose {CloSeR}, a staged framework that builds a domain-adapted closed-set teacher via efficient block-wise adapter tuning and transfers its knowledge to downstream GCD training.
\textbf{(iii)} {Within CloSeR}, we introduce {Unified Relational Distillation (URD)}, a head-agnostic relational regularizer that distills complementary global and local relations through separate feature pathways.
\textbf{(iv)} We validate CloSeR through extensive experiments across multiple datasets, backbones, and heterogeneous GCD baselines, showing consistent gains and strong generalization.

%% file: secs/2_related.tex
\section{Related Work}
\label{sec:related}

\noindent\textbf{Category Discovery.} 
Category discovery~\cite{he2025category}, first formulated as Novel Category Discovery (NCD) in a transfer-clustering setting~\cite{han2019learning}, seeks to transfer knowledge from labelled (seen) categories to cluster unlabelled data from novel classes. Generalized Category Discovery (GCD)~\cite{vaze2022generalized} relaxes the NCD assumption by allowing the unlabelled pool to contain samples from both known and unknown classes, prompting extensive follow-up work that explores diverse strategies for this challenging setting~\cite{han2020automatically, han2021autonovel, jia2021joint, zhao2021novel, ncl, fini2021unified, wu2022efficient, cao2022open,hao2023cipr,pu2023dynamic,joseph2022novel,cendra2024promptccd,wang2024hilo,liu2025debgcd,liu2025hyperbolic,he2025seal,cao2025allgcd,Cendra2026PartCo,tangbures,tang2026cogegcd,tang2025ocrt}. 
In the common GCD setup, a self-supervised pretrained image encoder is used as the backbone~\cite{caron2021emerging,oquab2023dinov2}; based on how category indices are predicted, methods are typically grouped into \textit{parametric} and \textit{non-parametric}.
On the \textit{parametric} side, SimGCD~\cite{wen2023parametric} first proposes to learn a parametric classifier regularized by mean-entropy minimization; SPTNet~\cite{wang2024sptnet} extends SimGCD by introducing spatial prompt tuning to better focus on discriminative object parts and strengthen knowledge transfer; DebGCD~\cite{liu2025debgcd} augments SimGCD~\cite{wen2023parametric} with a hard-label debiased auxiliary classifier and a semantic-distribution detector to identify shifts between known and unknown classes; 
AF~\cite{xu2025hidden} mitigates distracted attention in GCD by measuring token importance and adaptively pruning background/non-informative tokens;
APL~\cite{dai2025adaptive} replaces global features with adaptive part-based representations discovered via learnable queries to balance discriminability and generalization;
MOS~\cite{peng2025mos} treats scene context as a prior in GCD and models object--scene associations to resolve ambiguity between base/novel classes;
and SEAL~\cite{he2025seal} proposes a semantic-aware hierarchical framework that propagates coarse-to-fine supervision from labelled to unlabelled data via hierarchical contrastive learning and a cross-granularity consistency module. 
On the \textit{non-parametric} side, SelEx~\cite{RastegarECCV2024} employs hierarchical semi-supervised $k$-means and achieves strong results on fine-grained datasets, while HypCD~\cite{liu2025hyperbolic} operates in hyperbolic space to capture hierarchical class relations and offers a general GCD framework in that geometry.
More recently, ConGCD~\cite{tang2025dissecting} tackles GCD by learning primitive-oriented representations via semantic reconstruction and combining dominant/contextual consensus units with a scheduler to integrate complementary cues for recognizing both known and novel categories.

\noindent\textbf{Knowledge Distillation.}
Knowledge Distillation (KD)~\cite{gou2021knowledge} transfers knowledge from a high-capacity teacher to a compact student and has been widely applied across computer vision, including image classification~\cite{muller2019does,duan2025deep,wang2026gorag}, object detection~\cite{chen2017learning,wang2019distilling,liu2020multiple,shen2021distilled,yang2021r3det,cheng2024uadet}, and semantic segmentation~\cite{liu2019structured,he2019knowledge}.
The classical formulation~\cite{hinton2015distilling} matches softened logits to convey ``dark knowledge'' beyond hard labels. Extensions exploit intermediate representations: FitNets~\cite{romero2014fitnets} use intermediate ``hints'', and attention transfer~\cite{zagoruyko2017paying} guides students via spatial attention maps.
Similarity- and relation-based KD align embedding structures rather than only logits, including similarity-preserving KD~\cite{tung2019similarity}, correlation congruence~\cite{peng2019correlation}, relational KD~\cite{park2019relational}, and contrastive KD~\cite{tian2019contrastive}.
Apart from teacher--student pairs of different networks, self-distillation replaces the external teacher with a model's own multi-branch or temporal outputs~\cite{mobahi2020self,yun2020regularizing}, while momentum teachers stabilize targets in semi/self-supervised regimes~\cite{tarvainen2017mean,grill2020bootstrap,caron2021emerging,zhou2021image,assran2022masked}. Teacher--student asymmetry is central to self-supervised learning, underpinning BYOL~\cite{grill2020bootstrap}, DINO~\cite{caron2021emerging}, iBOT~\cite{zhou2021image}, and masked hybrids~\cite{assran2022masked}.
Within GCD, self-distillation techniques~\cite{caron2021emerging,assran2022masked} have become a common foundation.

In this work, we propose to inject the categorical prior from the closed-set models into the GCD models by distilling the relations. Unlike previous relational KD~\cite{park2019relational}, we consider both \emph{global} and \emph{local} relations in two \emph{decoupled} feature spaces.
Our framework unifies distillation to heterogeneous student models by transferring teacher-induced structure, allowing both \textit{parametric} and \textit{non-parametric} GCD methods with a single training objective.

%% file: secs/3_method.tex
\section{Preliminaries}

\label{sec:preliminaries}
\noindent\textbf{Problem Setup and Notation.} 
GCD~\cite{vaze2022generalized} targets a model that jointly handles two goals: classify unlabelled samples belonging to known classes and partition the remaining unlabelled samples into clusters of unknown classes. Let $\mathbf{D}_u = \{(\mathbf{x}^{u}_{i}, {y}^{u}_{i})\} \subset \mathbf{X} \times \mathbf{Y}_u$ be the unlabelled set and $\mathbf{D}_l = \{(\mathbf{x}^{l}_{i}, {y}^{l}_{i})\} \subset \mathbf{X} \times \mathbf{Y}_l$ be the labelled set, with $\mathbf{Y}_u$ and $\mathbf{Y}_l$ denoting their label spaces. 
By construction, the unlabelled pool spans both known and novel categories, and in particular $\mathbf{Y}_l \subset \mathbf{Y}_u$. We write $M=|\mathbf{Y}_l|$ for the number of labelled classes. Consistent with prior work~\cite{han2021autonovel,wen2023parametric,vaze2023no}, we assume the total number of categories $K=|\mathbf{Y}_l \cup \mathbf{Y}_u|$ is given; when $K$ is unknown, it can be estimated using existing techniques~\cite{han2019learning,vaze2022generalized}.

\noindent\textbf{Baselines.}
Vaze \etal~\cite{vaze2022generalized} formalize the task and introduce an early \textit{non-parametric} baseline.
The method fine-tunes a self-supervised foundation model~\cite{caron2021emerging} to improve representation quality by jointly optimizing a supervised contrastive objective on labelled data and a self-supervised contrastive objective on all data.
Concretely, given two random augmentations $\mathbf{x}_i$ and $\mathbf{x}_i'$ of the same image within a mini-batch ${B}$, the self-supervised contrastive loss is:
\begin{equation}
\textstyle
    \mathcal{L}_\text{rep}^{u} = \frac{1}{|{B}|}\sum_{i\in {B}} -\log\frac{\exp(\mathbf{z}_i \cdot \mathbf{z}_i'/\tau_r)}{\sum\nolimits_{j\neq i}\exp(\mathbf{z}_i\cdot \mathbf{z}_j'/\tau_r)},
\end{equation}
where $\mathbf{z}_i=\operatorname{L2Norm}(\mathcal{M}_{\text{\tiny GCD}}(f_{\theta}(\mathbf{x}_i)))$ denotes the $\ell_2$-normalized projected feature, $\mathbf{z}_i'$ is the feature from the other view $\mathbf{x}_i'$, $f_{\theta}$ is the backbone, $\mathcal{M}_{\text{\tiny GCD}}$ is the projection head, and $\tau_r$ is the temperature.
The supervised contrastive loss is defined as:
\begin{equation}
\textstyle
    \mathcal{L}_\text{rep}^{s} = \frac{1}{|{B}_l|}\sum_{i\in {B}_l} \frac{1}{|{N}_i|}\sum\limits_{q\in {N}_i}-\log\frac{\exp(\mathbf{z}_i \cdot \mathbf{z}_q/\tau_r)}{\sum\nolimits_{j\neq i}\exp(\mathbf{z}_i \cdot \mathbf{z}_j/\tau_r)},
\end{equation}
where ${N}_i$ indexes the samples in the labelled mini-batch ${B}_l \subset {B}$ that share the same label as $\mathbf{x}_i$.
The overall representation objective is $\mathcal{L}_\text{rep}=(1-\lambda)\mathcal{L}_\text{rep}^{u} + \lambda\mathcal{L}_\text{rep}^{s}$, where $\lambda$ is the balance factor. Building on this line of work, Rastegar \etal~\cite{RastegarECCV2024} propose SelEx, a stronger hierarchical \textit{non-parametric} method for fine-grained GCD, which has become a widely used baseline in the field.

Wen \etal~\cite{wen2023parametric} propose SimGCD, the first \textit{parametric} GCD baseline that has been widely adopted in subsequent work~\cite{vaze2023no,wang2024sptnet}.
It learns a parametric classifier via a self-distillation framework~\cite{caron2021emerging}, initialized with $K$ normalized prototypes $\mathbf{C}=\{\mathbf{c}_1,\dots,\mathbf{c}_K\}$.
For an augmented view $\mathbf{x}_i$ with normalized hidden feature $\mathbf{h}_i=f_\theta(\mathbf{x}_i)/||f_\theta(\mathbf{x}_i)||_2$, the predicted probability of class $k$ is:
\begin{equation}
\textstyle
    {\mathbf{p}_i^{(k)}} = \frac{\exp(\mathbf{h}_i\cdot\mathbf{c}_k/\tau_s)}{\sum\nolimits_{j=1}^K \exp(\mathbf{h}_i\cdot\mathbf{c}_j/\tau_s)},
\end{equation}
where $\tau_s$ is the student temperature.
A soft target $\mathbf{q}_i$ is obtained from the teacher prediction computed on another augmented view using a sharper temperature $\tau_t$.
The self-distillation objective is the cross-entropy between the two views,
$\ell_{\text{ce}}(\mathbf{q}_i',\mathbf{p}_i)=-\sum\nolimits_{j=1}^K \mathbf{q}_i'^{(j)}\log \mathbf{p}_i^{(j)}$.
Accordingly, the unsupervised loss over the mini-batch $B$ is:
\begin{equation}
\textstyle
    \mathcal{L}_\text{cls}^{u}=\frac{1}{|{B}|}\sum_{i\in {B}} \ell_{\text{ce}}(\mathbf{q}'_i,\mathbf{p}_i)-\xi \mathcal{H}(\overline{\mathbf{p}}),
\end{equation}
where $\overline{\mathbf{p}}=\frac{1}{2|{B}|}\sum\nolimits_{i\in {B}}(\mathbf{p}_i+\mathbf{p}'_i)$ denotes the batch-mean prediction and $\mathcal{H}(\overline{\mathbf{p}})=-\sum\nolimits_{j=1}^K \overline{\mathbf{p}}^{(j)}\log \overline{\mathbf{p}}^{(j)}$ is its entropy regularizer, weighted by $\xi$.
For labelled samples, the supervised classification loss is
$\mathcal{L}_\text{cls}^s=\frac{1}{|{B}_l|}\sum\nolimits_{i\in {B}_l}\ell_{\text{ce}}(\mathbf{p}_i,\mathbf{y}_i)$, where $\mathbf{y}_i$ is the one-hot encoding of $y_i$.
The overall classification objective is $\mathcal{L}_\text{cls}=(1-\lambda)\mathcal{L}_\text{cls}^u+\lambda\mathcal{L}_\text{cls}^s$, with $\lambda$ being the balance factor.
Together with the representation objective $\mathcal{L}_\text{rep}$, the full training loss is $\mathcal{L}_{\text{\tiny GCD}} = \mathcal{L}_\text{cls} + \mathcal{L}_\text{rep}$.

\noindent\textbf{Limitation of Baselines.}
All the aforementioned baselines~\cite{vaze2022generalized,wen2023parametric,RastegarECCV2024} follow a \emph{one-stage} paradigm, directly fine-tuning the backbone on mixed labelled--unlabelled data by jointly optimizing supervised and unsupervised objectives (\eg, $\mathcal{L}_\text{rep}^{s}$ with $\mathcal{L}_\text{rep}^{u}$, and $\mathcal{L}_\text{cls}^{s}$ with $\mathcal{L}_\text{cls}^{u}$).
This coupling entangles known-class recognition with novel-class discovery: unsupervised signals computed over all samples (often from noisy pseudo-labels) can conflict with closed-set discrimination and bias representations toward known categories.
Moreover, updating pretrained features under mixed objectives can distort the foundation model geometry---especially with limited labelled data---leading to suboptimal \textit{transfer clustering} for unknown classes.

\begin{figure*}[t]
    \centering
    \includegraphics[width=1.0\textwidth]{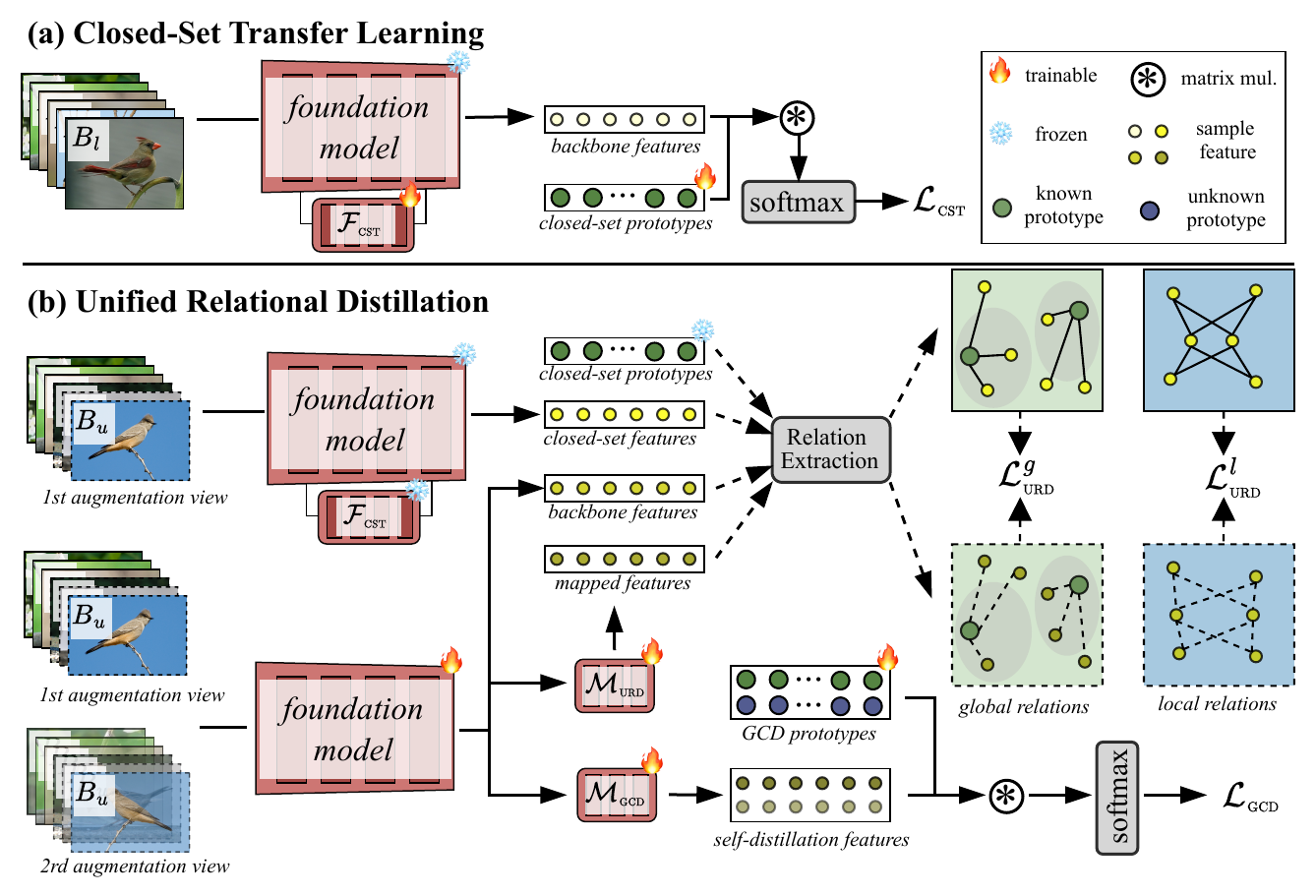}
    \caption{Overall pipeline of CloSeR (illustrated with SimGCD as an example; it is also applicable to other baselines): \textbf{(a)} Closed-Set Transfer Learning; \textbf{(b)} Unified Relational Distillation. \emph{Note}: The representation loss in SimGCD is omitted for clarity.
    }
    \label{fig:method}
\end{figure*}

\section{Method}
\label{sec:method}
\subsection{Overview}
We propose \textbf{CloSeR}, a concise, staged framework with two components (Fig.~\ref{fig:method}). 
\textbf{First}, we perform {Closed-Set Transfer Learning (CST)} by inserting a block-wise adapter~\cite{chen2022adaptformer} into a \textit{frozen} foundation model (\eg, DINO/DINOv2) and learning $M$ closed-set prototypes using labelled data only.
\textbf{Second}, during GCD training we introduce {Unified Relational Distillation (URD)}, which distills \textbf{(i)} global sample-to-prototype relations and \textbf{(ii)} local sample-to-sample relations from the closed-set space into the GCD student. We train the student with the standard objective (Sec.~\ref{sec:preliminaries}) together with our URD regularization.

\subsection{Closed-Set Transfer Learning}
\label{sec:method:CST}
Let $\phi$ denote a pretrained ViT foundation model (\eg, DINO~\cite{caron2021emerging}/DINOv2~\cite{oquab2023dinov2}) trained on large-scale data. 
Directly fine-tuning all parameters of $\phi$ on the limited labelled set $\mathbf{D}_l$ is often inefficient and may overfit, while keeping $\phi$ fully frozen can leave a non-negligible domain gap between the pre-training distribution and the target dataset.
To obtain an effective yet stable transfer, we insert a lightweight, \emph{block-wise} adapter~\cite{chen2022adaptformer,jia2022vpt,Zhan2025ELIP} into each transformer block.
This design enables \emph{feature adaptation throughout the network hierarchy}---from early blocks capturing low-level appearance cues to late blocks encoding high-level semantics---while preserving the strong generalization of the original foundation weights.
We denote the collection of all adapter parameters by $\mathcal{F}_{\text{\tiny CST}}$ and freeze all original parameters of $\phi$.

Given a randomly augmented view $\mathbf{x}_i$ of an image, the closed-set transfer model outputs an $\ell_2$-normalized feature:
\begin{equation}
\mathbf{h}^\text{cs}_i
=
\frac{\phi(\mathbf{x}_i;\mathcal{F}_{\text{\tiny CST}})}
{\|\phi(\mathbf{x}_i;\mathcal{F}_{\text{\tiny CST}})\|_2}.
\end{equation}
We use normalized features for two reasons. First, this matches the cosine-similarity classifier used in both the GCD baseline (Sec.~\ref{sec:preliminaries}) and the foundation model training~\cite{caron2021emerging,oquab2023dinov2}, where logits are computed by inner products between normalized features and normalized prototypes and scaled by a temperature. Second, it stabilizes optimization when transferring from a foundation embedding space to the target domain and makes the learned prototypes comparable across classes.

We maintain $M$ $\ell_2$-normalized closed-set prototypes $\mathbf{C}^\text{cs}=\{\mathbf{c}^\text{cs}_1,\dots,\mathbf{c}^\text{cs}_M\}$, one per known class, acting as a cosine classifier head in the adapted closed-set space.
For a labelled sample $(\mathbf{x}^l_i,y^l_i)$, we compute the probability of class $k$ as:
\begin{equation}
{\mathbf{p}^\text{cs}_i}^{(k)}
=
\frac{\exp(\mathbf{h}^\text{cs}_i\cdot \mathbf{c}^\text{cs}_k/\tau_\text{cs})}
{\sum_{j=1}^{M}\exp(\mathbf{h}^\text{cs}_i\cdot \mathbf{c}^\text{cs}_j/\tau_\text{cs})},
\end{equation}
where $\tau_\text{cs}$ is consistent with the temperature used in $\mathcal{L}_{\text{cls}}^s$.
We train CST with the supervised cross-entropy loss:
\begin{equation}
\mathcal{L}_{\text{\tiny CST}}
=
\frac{1}{|B_l|}
\sum_{i\in B_l}
\ell_{\text{ce}}(\mathbf{p}^\text{cs}_i,\mathbf{y}^l_i),
\end{equation}
where $\mathbf{y}^l_i$ is the one-hot encoding of $y^l_i$.
The resulting adapted closed-set representation and prototypes are then used to construct the teacher global/local relations distilled by URD (Sec.~\ref{sec:method:urd}).

\subsection{Unified Relational Distillation}
\label{sec:method:urd}
While the baseline optimizes a $K$-way prototype classifier with self-distillation techniques (Sec.~\ref{sec:preliminaries}), its discovery of novel categories can be unstable under limited supervision, and the learned embedding may drift away from the strong semantic structure provided by the foundation model.
We therefore introduce {Unified Relational Distillation (URD)}, which transfers \emph{relational} knowledge from the closed-set adapted foundation model into the GCD student model.
Compared to matching features directly, relation-based distillation~\cite{park2019relational} is more tolerant to representation mismatch and focuses on preserving the \emph{geometry} of the embedding space that has been proven to be critical for category discovery~\cite{he2025seal,liu2025hyperbolic}.

At each iteration, we first construct a mixed mini-batch $B = B_l \cup B_u$, where $B_l$ and $B_u$ are sampled from $\mathbf{D}_l$ and $\mathbf{D}_u$, respectively.
Following the standard practice, \emph{two} augmented views are used to compute the standard self-distillation losses; however, we empirically find that applying URD to only \emph{one} view per image is sufficient. 
We denote the selected augmented view by $\mathbf{x}_i$ for $i\in B$ and treat the closed-set adapted model as a teacher.
Specifically, we \textit{freeze} the foundation backbone $\phi$ and the learned adapters $\mathcal{F}_{\text{\tiny CST}}$, as well as the closed-set prototypes $\mathbf{C}^{\text{cs}}$, and compute normalized teacher features $\mathbf{h}^{T}_i = \mathbf{h}^\text{cs}_i$.
Let $f_\theta$ denote the student backbone, initialized from the same foundation model.
During GCD training, we fine-tune only the last few transformer blocks (and keep earlier blocks frozen) together with the GCD heads/prototypes.
For relational distillation, we extract normalized student features:
\begin{equation}
\mathbf{h}^{S}_i = \frac{f_\theta(\mathbf{x}_i)}{\|f_\theta(\mathbf{x}_i)\|_2},\quad i\in B.
\end{equation}

To compare sample-to-prototype relations in the closed-set space, we further introduce a lightweight mapping network $\mathcal{M}_{\text{\tiny URD}}$ and define mapped features:
\begin{equation}
\tilde{\mathbf{h}}^{S}_i=\frac{\mathcal{M}_{\text{\tiny URD}}(\mathbf{h}^{S}_i)}{\|\mathcal{M}_{\text{\tiny URD}}(\mathbf{h}^{S}_i)\|_2}.
\end{equation}
This mapping compensates for the representational gap between the GCD student features and the closed-set adapted teacher space, enabling effective relation transfer without constraining the student to mimic teacher features directly.

After obtaining features from the teacher and student models, we define \emph{global relations} as the similarity-induced distribution over the $M$ closed-set prototypes for each sample. 
Using the fixed closed-set prototypes $\mathbf{C}^\text{cs}$, the element in the global relation matrix $\mathbf{R}^{T}_{g}$ of the teacher model is:
\begin{equation}
\textstyle
\mathbf{R}^{T}_{g}(i,k)=
\frac{\exp(\mathbf{h}^{T}_i\cdot \mathbf{c}^\text{cs}_k/\tau_g)}
{\sum_{j=1}^{M}\exp(\mathbf{h}^{T}_i\cdot \mathbf{c}^\text{cs}_j/\tau_g)},
\end{equation}
and the student counterpart is computed with mapped features:
\begin{equation}
\textstyle
\mathbf{R}^{S}_{g}(i,k)=
\frac{\exp(\tilde{\mathbf{h}}^{S}_i\cdot \mathbf{c}^\text{cs}_k/\tau_g)}
{\sum_{j=1}^{M}\exp(\tilde{\mathbf{h}}^{S}_i\cdot \mathbf{c}^\text{cs}_j/\tau_g)}.
\end{equation}
After obtaining the relation matrices, we align the student global relations to the teacher via KL divergence with soft targets:
\begin{equation}
\textstyle
\mathcal{L}^{g}_{\text{\tiny URD}}
=
\frac{1}{|B|}
\sum_{i\in B}
\text{KL}\!\left(\mathbf{R}^{T}_{g}(i,\cdot)\,\|\,\mathbf{R}^{S}_{g}(i,\cdot)\right).
\end{equation}
Intuitively, this loss encourages the student to preserve the teacher's relative similarity profile to closed-set anchors, which stabilizes the semantic organization of the embedding while still allowing the student to allocate capacity to novel categories.

Global relations anchor each sample to closed-set prototypes; in addition, we distill \emph{local relations} that describe neighborhood structure within the mixed batch.
For $i\neq j$, we compute temperature-scaled cosine similarities:
\begin{equation}
\textstyle
s^{T}_{ij}=\frac{\mathbf{h}^{T}_i\cdot \mathbf{h}^{T}_j}{\tau_\ell},
\qquad
s^{S}_{ij}=\frac{\mathbf{h}^{S}_i\cdot \mathbf{h}^{S}_j}{\tau_\ell}.
\end{equation}
We then convert similarities into row-stochastic distributions, and the elements of the local relation matrices $\mathbf{R}^{T}_{\ell}$/$\mathbf{R}^{S}_{\ell}$ are defined as:
\begin{equation}
\textstyle
\mathbf{R}^{T}_{\ell}(i,j)=\frac{\exp(s^{T}_{ij})}{\sum_{t\in B,\,t\neq i}\exp(s^{T}_{it})},
\qquad
\mathbf{R}^{S}_{\ell}(i,j)=\frac{\exp(s^{S}_{ij})}{\sum_{t\in B,\,t\neq i}\exp(s^{S}_{it})}.
\end{equation}
These distributions capture, for each anchor sample, which other samples in the batch are most semantically related according to the teacher/student.
The neighborhood structures within the local relations of the two models are then matched via KL divergence:
\begin{equation}
\textstyle
\mathcal{L}^{l}_{\text{\tiny URD}}
=
\frac{1}{|B|}
\sum_{i\in B}
\mathrm{KL}\!\left(\mathbf{R}^{T}_{\ell}(i,\cdot)\,\|\,\mathbf{R}^{S}_{\ell}(i,\cdot)\right).
\end{equation}
This term encourages the student to preserve fine-grained relational geometry (\eg, intra-class compactness and inter-class separation) induced by the closed-set adapted teacher, which we find beneficial for both known-class recognition and novel-class clustering.
Finally, we optimize the standard objective together with URD:
\begin{equation}
\textstyle
\mathcal{L}_{\text{\tiny CloSeR}}
=
\mathcal{L}_{\text{\tiny GCD}}
+
\alpha\,\mathcal{L}^{g}_{\text{\tiny URD}}
+
\beta\,\mathcal{L}^{l}_{\text{\tiny URD}},
\end{equation}
where $\alpha$ and $\beta$ balance global and local relational transfer.

%% file: secs/4_experiment.tex
\section{Experiments}
\subsection{Implementation Details}
We benchmark {CloSeR} against two representative baselines: the \textit{non-parametric} SelEx~\cite{RastegarECCV2024} and the \textit{parametric} SimGCD~\cite{wen2023parametric}. Performance is evaluated using clustering accuracy (\textit{ACC})~\cite{vaze2022generalized}.
To study the effect of different foundation models, we instantiate all methods with pretrained weights from DINO~\cite{caron2021emerging} and DINOv2~\cite{oquab2023dinov2}. Unless stated otherwise, we use a ViT-B backbone for both teacher and student to ensure fair comparison across methods (ViT-B/14 for DINOv2 and ViT-B/16 for DINO).
In the first CST stage, we train the adapter for $100$ epochs on fine-grained datasets and $30$ epochs on generic datasets with $\tau_\text{cs}=0.1$, using an initial learning rate of $10^{-1}$ that is cosine-annealed to $10^{-4}$.
In the subsequent URD stage, for $\mathcal{L}_{\text{\tiny GCD}}$ we follow the hyperparameters and augmentation settings of the corresponding baseline. For $\mathcal{L}_{\text{\tiny URD}}$, we set the temperatures $\tau_g$ and $\tau_\ell$ to $0.3$, and the loss weights $\alpha$ and $\beta$ to $5.0$ and $20.0$, respectively.
Additional details on datasets, optimization, and hyperparameters are provided in the supplementary material.

\begin{table*}[t]
  \caption{Results on generic datasets (CIFAR-10/100~\cite{krizhevsky2009learning} and ImageNet-100~\cite{deng2009imagenet}), reported for \textit{All}/\textit{Old}/\textit{New} categories. Right: average over datasets.}
  \label{tab:results_generic}
  \centering
  \setlength{\tabcolsep}{1.5mm}{
  \resizebox{1.0\linewidth}{!}{
\begin{tabular}{@{}l@{}r@{}l>{\columncolor{my_blue}}ccc>{\columncolor{my_blue}}ccc>{\columncolor{my_blue}}ccc>{\columncolor{my_blue}}cccc@{}}
\toprule
&&&\multicolumn{3}{c}{CIFAR-10~\cite{krizhevsky2009learning}}&
\multicolumn{3}{c}{CIFAR-100~\cite{krizhevsky2009learning}}&
\multicolumn{3}{c}{ImageNet-100~\cite{deng2009imagenet}}&
\multicolumn{3}{c}{Average}\\
\cmidrule(lr){4-6} \cmidrule(lr){7-9} \cmidrule(lr){10-12} \cmidrule(lr){13-15}
&Method&\pub{Venue}&
All&Old&New&All&Old&New&All&Old&New&
All&Old&New\\
\midrule
\multirow{13}{*}{\rotatebox{90}{\emph{DINO}}}
&GCD~\cite{vaze2022generalized}&\pub{CVPR 2022}
&91.5&{97.9}&88.2 &73.0&76.2&66.5 &74.1&89.8&66.3
&\avgthree{91.5}{73.0}{74.1}&\avgthree{97.9}{76.2}{89.8}&\avgthree{88.2}{66.5}{66.3}\\
&PromptCAL~\cite{zhang2023promptcal}&\pub{CVPR 2023}
&\textbf{97.9}&96.6&{98.5} &81.2&84.2&75.3 &83.1&92.7&78.3
&\avgthree{97.9}{81.2}{83.1}&\avgthree{96.6}{84.2}{92.7}&\avgthree{98.5}{75.3}{78.3}\\
&GPC~\cite{Zhao_2023_ICCV}&\pub{ICCV 2023}
&90.6&97.6&87.0 &75.4&{84.6}&60.1 &75.3&93.4&66.7
&\avgthree{90.6}{75.4}{75.3}&\avgthree{97.6}{84.6}{93.4}&\avgthree{87.0}{60.1}{66.7}\\
&InfoSieve~\cite{rastegar2023learn}&\pub{NeurIPS 2023}
&94.8&97.7&93.4 &78.3&82.2&70.5 &80.5&93.8&73.8
&\avgthree{94.8}{78.3}{80.5}&\avgthree{97.7}{82.2}{93.8}&\avgthree{93.4}{70.5}{73.8}\\
&SPTNet~\cite{wang2024sptnet}&\pub{ICLR 2024}
&{97.3}&95.0&\underline{98.6} &81.3&84.3&75.6 &{85.4}&93.2&{81.4}
&\avgthree{97.3}{81.3}{85.4}&\avgthree{95.0}{84.3}{93.2}&\avgthree{98.6}{75.6}{81.4}\\
&DebGCD~\cite{liu2025debgcd}&\pub{ICLR 2025}
&97.2&94.8&98.4 &83.0&84.6&79.9 &\underline{85.9}&\underline{94.3}&\underline{81.6}
&88.7&91.2&\underline{86.6}\\
&HypCD~\cite{liu2025hyperbolic}&\pub{CVPR 2025}
&96.7&97.6&96.3 &{82.4}&85.1&77.0 &\textbf{86.8}&\textbf{94.6}&\textbf{82.8}
&88.6&\underline{92.4}&85.4\\
&AF~\cite{xu2025hidden}&\pub{ICCV 2025} 
&\underline{97.8}&95.9&\textbf{98.8} &82.2&85.0&76.5 &85.4&\textbf{94.6}&80.8
&\avgthree{97.8}{82.2}{85.4}&\avgthree{95.9}{85.0}{94.6}&\avgthree{98.8}{76.5}{80.8}\\
&SEAL~\cite{he2025seal}&\pub{NeurIPS 2025}
&97.2&94.7&98.4 &82.1&81.7&\textbf{83.0} &84.6&90.9&81.3
&88.0&89.1&\textbf{87.6}\\

\cmidrule(l{2mm}){2-15}

&SimGCD~\cite{wen2023parametric}&\pub{ICCV 2023}
&97.1&95.1&98.1 &80.1&81.2&77.8 &83.0&93.1&77.9
&\avgthree{97.1}{80.1}{83.0}&\avgthree{95.1}{81.2}{93.1}&\avgthree{98.1}{77.8}{77.9}\\
&\multicolumn{2}{r}{\textbf{+ CloSeR}}
&97.6&\textbf{99.3}&96.7 &\textbf{83.8}&\underline{85.3}&80.8	 &84.9&92.6&81.0
&\underline{88.8}&\underline{92.4}&86.2\\
&SelEx~\cite{RastegarECCV2024}&\pub{ECCV 2024}
&95.9&{98.1}&94.8 &{82.3}&{85.3}&76.3 &83.1&93.6&77.8
&\avgthree{95.9}{82.3}{83.1}&\avgthree{98.1}{85.3}{93.6}&\avgthree{94.8}{76.3}{77.8}\\
&\multicolumn{2}{r}{\textbf{+ CloSeR}}
&{97.7}&\underline{98.3}&97.3 &\underline{83.7}&\textbf{88.4}&74.3 &85.6&\textbf{94.6}&81.1
&\textbf{89.0}&\textbf{93.8}&84.2\\

\midrule

\multirow{10}{*}{\rotatebox{90}{\emph{DINOv2}}}
&GCD~\cite{vaze2022generalized}&\pub{CVPR 2022}
&97.8&\textbf{99.0}&97.1 &79.6&84.5&69.9 &78.5&89.5&73.0
&\avgthree{97.8}{79.6}{78.5}&\avgthree{99.0}{84.5}{89.5}&\avgthree{97.1}{69.9}{73.0}\\
&DebGCD~\cite{liu2025debgcd}&\pub{ICLR 2025}
&\textbf{98.9}&97.5&\underline{99.6} &\underline{90.1}&90.9&88.6 &\textbf{93.2}&\textbf{97.0}&\textbf{91.2}
&\underline{94.1}&95.1&\underline{93.1}\\
&$^\dagger$HypCD~\cite{liu2025hyperbolic}&\pub{CVPR 2025}
&98.6&98.1&98.9
&88.6&{91.5}&82.8
&{92.3}&{96.4}&{90.2}
&\avgthree{98.6}{88.6}{92.3}&\avgthree{98.1}{91.5}{96.4}&\avgthree{98.9}{82.8}{90.2}\\
&SEAL~\cite{he2025seal}&\pub{NeurIPS 2025}
&\textbf{98.9}&98.1&99.3 &89.8&90.4&\underline{89.5} &91.3&93.3&90.3
&93.3&93.9&{93.0}\\

\cmidrule(l{2mm}){2-15}

&SimGCD~\cite{wen2023parametric}&\pub{ICCV 2023}
&\underline{98.7}&96.7&{99.7} &88.5&{89.2}&{87.2} &89.9&95.5&87.1
&\avgthree{98.7}{88.5}{89.9}&\avgthree{96.7}{89.2}{95.5}&\avgthree{99.7}{87.2}{87.1}\\
&\multicolumn{2}{r}{\textbf{+ CloSeR}}
&\textbf{98.9}&97.1&\textbf{99.8} &\textbf{93.2}&\textbf{93.0}&\textbf{93.5}	 &92.5&\underline{96.6}&90.4
&\textbf{94.9}&\underline{95.6}&\textbf{94.6} \\
&SelEx~\cite{RastegarECCV2024}&\pub{ECCV 2024}
&98.5*&98.8*&98.5* &87.7*&{90.8}*&81.5* &90.9*&{96.2}*&88.3*
&\avgthree{98.5}{87.7}{90.9}&\avgthree{98.8}{90.8}{96.2}&\avgthree{98.5}{81.5}{88.3}\\
&\multicolumn{2}{r}{\textbf{+ CloSeR}}
&\underline{98.7}&\underline{98.9}&98.6 &89.7&\underline{92.7}&83.7 &\underline{92.7}&95.7&\underline{91.1}
&93.7&\textbf{95.8}&91.1\\

\bottomrule
\end{tabular}
}} 
\RaggedRight
\noindent\scriptsize{$^\dagger$results based on the best baseline~\cite{RastegarECCV2024}. *results from~\cite{liu2025hyperbolic}.}
\end{table*}

\begin{table*}[t]
  \caption{Results on fine-grained datasets (CUB~\cite{wah2011caltech}, Stanford-Cars~\cite{krause20133d}, FGVC-Aircraft~\cite{maji2013fine}), reported for \textit{All}/\textit{Old}/\textit{New} categories. Right: average over datasets.}
  \label{tab:results_finegrained}
  \centering
  \setlength{\tabcolsep}{1.5mm}{
  \resizebox{1.0\linewidth}{!}{
\begin{tabular}{@{}l@{}r@{}l>{\columncolor{my_blue}}ccc>{\columncolor{my_blue}}ccc>{\columncolor{my_blue}}ccc>{\columncolor{my_blue}}ccc@{}}
\toprule
&&&\multicolumn{3}{c}{CUB~\cite{wah2011caltech}}&
\multicolumn{3}{c}{Stanford-Cars~\cite{krause20133d}}&
\multicolumn{3}{c}{FGVC-Aircraft~\cite{maji2013fine}}&
\multicolumn{3}{c}{Average}\\
\cmidrule(lr){4-6} \cmidrule(lr){7-9} \cmidrule(lr){10-12} \cmidrule(lr){13-15}
&Method&\pub{Venue}&
All&Old&New&
All&Old&New&
All&Old&New&
All&Old&New\\
\midrule
\multirow{16}{*}{\rotatebox{90}{\emph{DINO}}}
&GCD~\cite{vaze2022generalized}&\pub{CVPR 2022}
&51.3&56.6&48.7 &39.0&57.6&29.9 &45.0&41.1&46.9
&\avgthree{51.3}{39.0}{45.0}&\avgthree{56.6}{57.6}{41.1}&\avgthree{48.7}{29.9}{46.9}\\
&PromptCAL~\cite{zhang2023promptcal}&\pub{CVPR 2023}
&62.9&64.4&62.1 &50.2&70.1&40.6 &52.2&52.2&52.3
&\avgthree{62.9}{50.2}{52.2}&\avgthree{64.4}{70.1}{52.2}&\avgthree{62.1}{40.6}{52.3}\\
&GPC~\cite{Zhao_2023_ICCV}&\pub{ICCV 2023}
&52.0&55.5&47.5 &38.2&58.9&27.4 &43.3&40.7&44.8
&\avgthree{52.0}{38.2}{43.3}&\avgthree{55.5}{58.9}{40.7}&\avgthree{47.5}{27.4}{44.8}\\
&$\mu$GCD~\cite{vaze2023no}&\pub{NeurIPS 2023}
&65.7&68.0&64.6 &56.5&68.1&50.9 &53.8&55.4&53.0
&\avgthree{65.7}{56.5}{53.8}&\avgthree{68.0}{68.1}{55.4}&\avgthree{64.6}{50.9}{53.0}\\
&InfoSieve~\cite{rastegar2023learn}&\pub{NeurIPS 2023}
&69.4&{77.9}&65.2 &55.7&74.8&46.4 &56.3&63.7&52.5
&\avgthree{69.4}{55.7}{56.3}&\avgthree{77.9}{74.8}{63.7}&\avgthree{65.2}{46.4}{52.5}\\
&SPTNet~\cite{wang2024sptnet}&\pub{ICLR 2024}
&65.8&68.8&65.1 &59.0&{79.2}&49.3 &{59.3}&61.8&58.1
&\avgthree{65.8}{59.0}{59.3}&\avgthree{68.8}{79.2}{61.8}&\avgthree{65.1}{49.3}{58.1}\\
&FlipClass~\cite{lin2024flipped}&\pub{NeurIPS 2024}
&71.3&71.3&71.3 &63.1&\underline{81.7}&53.8 &59.3&66.9&55.4
&\avgthree{71.3}{63.1}{59.3}&\avgthree{71.3}{81.7}{66.9}&\avgthree{71.3}{53.8}{55.4}\\
&DebGCD~\cite{liu2025debgcd}&\pub{ICLR 2025}
&{66.3}&71.8&{63.5} &65.3&{81.6}&57.4 &{61.7}&63.9&{60.6}
&\avgthree{66.3}{65.3}{61.7}&\avgthree{71.8}{81.6}{63.9}&\avgthree{63.5}{57.4}{60.6}\\
&MOS~\cite{peng2025mos}&\pub{CVPR 2025}
&69.6&72.3&68.2&64.6&80.9&56.7&61.1&66.9&58.2
&\avgthree{69.6}{64.6}{61.1}&\avgthree{72.3}{80.9}{66.9}&\avgthree{68.2}{56.7}{58.2}\\
&APL~\cite{dai2025adaptive}&\pub{CVPR 2025}
&68.5&73.1&66.2 &62.3&80.7&53.4 &60.9&63.5&59.6
&\avgthree{68.5}{62.3}{60.9}&\avgthree{73.1}{80.7}{63.5}&\avgthree{66.2}{53.4}{59.6}\\
&HypCD~\cite{liu2025hyperbolic}&\pub{CVPR 2025}
&{79.8}&75.8&{81.8} &{62.9}&{80.0}&{54.7} &\underline{65.9}&{67.3}&\underline{65.1}
&\underline{69.5}&74.4&\underline{67.2}\\
&AF~\cite{xu2025hidden}&\pub{ICCV 2025} 
&69.0&74.3&66.3 &\textbf{67.0}&80.7&\textbf{60.4} &59.4&\underline{68.1}&55.0
&\avgthree{69.0}{67.0}{59.4}&\avgthree{74.3}{80.7}{68.1}&\avgthree{66.3}{60.4}{55.0}\\
&ConGCD~\cite{tang2025dissecting}&\pub{ICCV 2025}
&\underline{81.7}&\textbf{80.4}&\underline{82.4} &57.5&77.5&47.9 &62.5&\textbf{70.2}&58.7
&67.2&\underline{76.0}&63.0\\
&SEAL~\cite{he2025seal}&\pub{NeurIPS 2025}
&66.2&{72.1}&63.2&{65.3}&{79.3}&\underline{58.5}&{62.0}&{65.3}&{60.4}
&\avgthree{66.2}{65.3}{62.0}&\avgthree{72.1}{79.3}{65.3}&\avgthree{63.2}{58.5}{60.4}\\

\cmidrule(l{2mm}){2-15}

&SimGCD~\cite{wen2023parametric}&\pub{ICCV 2023}
&60.3&65.6&57.7 &53.8&71.9&45.0 &54.2&59.1&51.8
&\avgthree{60.3}{53.8}{54.2}&\avgthree{65.6}{71.9}{59.1}&\avgthree{57.7}{45.0}{51.8}\\
&\multicolumn{2}{r}{\textbf{+ CloSeR}}
&68.2&76.9&63.9 &63.0&80.2&54.7 &56.6&55.5&57.2
&\avgthree{68.2}{63.0}{56.6}&\avgthree{76.9}{80.2}{55.5}&\avgthree{63.9}{54.7}{57.2}\\
&SelEx~\cite{RastegarECCV2024}&\pub{ECCV 2024}
&{73.6}&75.3&{72.8} &58.5&75.6&50.3 &57.1&64.7&53.3
&\avgthree{73.6}{58.5}{57.1}&\avgthree{75.3}{75.6}{64.7}&\avgthree{72.8}{50.3}{53.3}\\
&\multicolumn{2}{r}{\textbf{+ CloSeR}}
&\textbf{82.8}&\underline{80.3}&\textbf{84.1} &\underline{66.5}&\textbf{83.4}&58.3 &\textbf{66.6}&\underline{68.1}&\textbf{65.8}
&\textbf{72.0}&\textbf{77.3}&\textbf{69.4}\\

\midrule
\multirow{10}{*}{\rotatebox{90}{\emph{DINOv2}}}
&GCD~\cite{vaze2022generalized}&\pub{CVPR 2022}
&71.9&71.2&72.3 &65.7&67.8&64.7 &55.4&47.9&59.2
&\avgthree{71.9}{65.7}{55.4}&\avgthree{71.2}{67.8}{47.9}&\avgthree{72.3}{64.7}{59.2}\\
&DebGCD~\cite{liu2025debgcd}&\pub{ICLR 2025}
&77.5&{80.8}&75.8 &75.4&87.7&{69.5} &{71.9}&{76.0}&{69.8}
&\avgthree{77.5}{75.4}{71.9}&\avgthree{80.8}{87.7}{76.0}&\avgthree{75.8}{69.5}{69.8}\\
&APL~\cite{dai2025adaptive}&\pub{CVPR 2025}
&75.1&79.1&73.2 &73.4&87.6&66.7 &68.8&74.1&66.6
&\avgthree{75.1}{73.4}{68.8}&\avgthree{79.1}{87.6}{74.1}&\avgthree{73.2}{66.7}{66.6}\\
&$^\dagger${HypCD}~\cite{liu2025hyperbolic}&\pub{CVPR 2025}
&\textbf{90.7}&85.3&\textbf{93.4} &\textbf{83.8}&{93.3}&\textbf{79.2} &\textbf{83.4}&{82.0}&\textbf{84.1}
&\textbf{86.0}&86.9&\textbf{85.6}\\
&$^\dagger$ConGCD~\cite{tang2025dissecting}&\pub{ICCV 2025}
&86.3&\textbf{87.4}&85.8 &79.8&93.1&73.3 &81.7&\textbf{83.3}&81.0
&82.6&\textbf{87.9}&80.0\\
&SEAL~\cite{he2025seal}&\pub{NeurIPS 2025}
&76.7&78.3&{75.9} &{77.7}&{88.7}&{72.4} &{74.6}&{73.2}&{75.3}
&\avgthree{76.7}{77.7}{74.6}&\avgthree{78.3}{88.7}{73.2}&\avgthree{75.9}{72.4}{75.3}\\

\cmidrule(l{2mm}){2-15}

&SimGCD~\cite{wen2023parametric}&\pub{ICCV 2023}
&71.5&78.1&68.3 &71.5&81.9&66.6 &63.9&69.9&60.9
&\avgthree{71.5}{71.5}{63.9}&\avgthree{78.1}{81.9}{69.9}&\avgthree{68.3}{66.6}{60.9}\\
&\multicolumn{2}{r}{\textbf{+ CloSeR}}
&78.1&79.5&77.3 &78.8&90.7&73.0 &71.8&68.7&73.3
&\avgthree{78.1}{78.8}{71.8}&\avgthree{79.5}{90.7}{68.7}&\avgthree{77.3}{73.0}{73.3}\\
&SelEx~\cite{RastegarECCV2024}&\pub{ECCV 2024}
&{87.4}&{85.1}&{88.5} &82.2&\underline{93.7}&76.7 &{79.8}&\underline{82.3}&78.6
&\avgthree{87.4}{82.2}{79.8}&\avgthree{85.1}{93.7}{82.3}&\avgthree{88.5}{76.7}{78.6}\\
&\multicolumn{2}{r}{\textbf{+CloSeR}}
&\underline{89.8}&\underline{86.5}&\underline{91.5} &\underline{83.1}&\textbf{94.6}&\underline{77.6} &\underline{82.1}&81.2&\underline{82.6}
&\underline{85.0}&\underline{87.4}&\underline{83.9}\\
\bottomrule
\end{tabular}
}}
\RaggedRight
{\scriptsize\noindent $^\dagger$results based on the best baseline~\cite{RastegarECCV2024}. \par}
\end{table*}

\subsection{Quantitative Comparison} 
Tab.~\ref{tab:results_generic} and Tab.~\ref{tab:results_finegrained} report results on three generic (CIFAR-10/100, ImageNet-100) datasets and three fine-grained (CUB, Stanford-Cars, FGVC-Aircraft) with DINO and DINOv2 backbones. 
For brevity, we sometimes abbreviate Stanford-Cars and FGVC-Aircraft as \textit{Cars} and \textit{Aircraft}, respectively.

\noindent\textbf{Generic Datasets.}
On generic benchmarks, the baselines are already strong, but CloSeR still provides consistent gains.
Under DINO, CloSeR improves the SimGCD baseline from 86.7 to 88.8 average \textit{All} accuracy (+2.1 points) and improves SelEx from 87.1 to 89.0 (+1.9 points).
Under DINOv2, SimGCD{+}CloSeR achieves 94.9 average \textit{All} and 94.6 average \textit{New} accuracy, delivering the \emph{best} reported average \textit{New} accuracy.
SelEx{+}CloSeR also improves the average \textit{All} accuracy (92.4$\rightarrow$93.7) with clear gains on CIFAR-100 and ImageNet-100.

\noindent\textbf{Fine-Grained Datasets.}
On fine-grained datasets, our CloSeR is particularly effective.
With DINO, CloSeR boosts SimGCD on CUB and Cars by large margins (\eg, Cars \textit{All}: 53.8$\rightarrow$63.0, +9.2 points), and improves its average \textit{All} accuracy from 56.1 to 62.6 ({+11.6\%} relative).
For SelEx, CloSeR brings even larger improvements: the average \textit{New} accuracy increases from 58.8 to 69.4 ({+18.0\%} relative) and the average \textit{All} increases from 63.1 to 72.0, yielding the \emph{best} DINO-based average performance.
With DINOv2, CloSeR continues to substantially improve SimGCD, raising average \textit{New} from 65.3 to 74.5 (+9.2 points; {+14.1\%} relative).
Meanwhile, SelEx{+}CloSeR reaches 85.0 average \textit{All} accuracy, which is the \emph{second-best} among DINOv2-based methods, and improves \textit{New} category discovery on all three datasets (\eg, CUB \textit{New}: 88.5$\rightarrow$91.5).

\noindent\textbf{Discussion.}
Across all six datasets and both backbones, CloSeR provides a robust performance boost when applied to heterogeneous GCD pipelines, demonstrating strong generality. The consistent improvements on both \textit{Old} and \textit{New} categories suggest that CloSeR mitigates the common training bias of existing methods and better preserves transferable structure from the foundation model.

\begin{table}[t]
\centering
\caption{Experimental results using different transfer learning methods with DINO~\cite{caron2021emerging} and DINOv2~\cite{oquab2023dinov2} pretrained weights on SimGCD. Results are reported on SSB~\cite{vaze2022semantic}.}
\setlength{\tabcolsep}{1.5mm}{
\resizebox{0.95\textwidth}{!}{
\begin{tabular}{lcc>{\columncolor{my_blue}}ccc>{\columncolor{my_blue}}ccc>{\columncolor{my_blue}}ccc}
        \toprule
        &&&\multicolumn{3}{c}{CUB~\cite{wah2011caltech}}&
        \multicolumn{3}{c}{Stanford-Cars~\cite{krause20133d}}&
        \multicolumn{3}{c}{FGVC-Aircraft~\cite{maji2013fine}}\\
        \cmidrule(lr{1em}){4-6} \cmidrule(lr{1em}){7-9} \cmidrule(lr{1em}){10-12}
        Adaptation Method&GCD Method&Pretrained &All&Old&New&All&Old&New&All&Old&New\\
        \midrule
        (1) Full Fine-tuning&SimGCD&DINO &68.1&76.3&\textbf{64.0} &60.9&77.5&52.9 &{54.8}&{53.2}&{55.6}\\
        (2) Fine-tuning Last Block&SimGCD&DINO &58.5&56.2&59.6 &46.7&56.8&40.3 &45.8&41.8&47.8\\
        (3) Block-wise Adapter&SimGCD&DINO &\textbf{68.2}&\textbf{76.9}&63.9 &\textbf{63.0}&\textbf{80.2}&\textbf{54.7} &\textbf{56.6}&\textbf{55.5}&\textbf{57.2}\\
                
        \midrule
        (1) Full Fine-tuning&SimGCD&DINOv2 &76.9&77.4&76.6 &78.4&91.7&72.0 &67.9&62.6	&70.5\\
        (2) Fine-tuning Last Block&SimGCD&DINOv2 &77.3&80.2&75.8 &77.9&90.5&71.8 &67.7&64.6&69.2\\
        (3) Block-wise Adapter&SimGCD&DINOv2 &\textbf{78.1}&\textbf{79.5}&\textbf{77.3} &\textbf{78.8}&\textbf{90.7}&\textbf{73.0} &\textbf{71.8}&\textbf{68.7}&\textbf{73.3}\\
        \bottomrule
\end{tabular}
}
}
\label{tab:adaptation}
\end{table}

\begin{table}[t]
\centering
\caption{Experimental results using different distillation components with DINO~\cite{caron2021emerging} pretrained weights on SelEx. Results are reported on the SSB~\cite{vaze2022semantic} benchmark.}
\setlength{\tabcolsep}{1.5mm}{
\resizebox{0.95\textwidth}{!}{
\begin{tabular}{ccc>{\columncolor{my_blue}}ccc>{\columncolor{my_blue}}ccc>{\columncolor{my_blue}}ccc}
        \toprule
        &&&\multicolumn{3}{c}{CUB~\cite{wah2011caltech}}&
        \multicolumn{3}{c}{Stanford-Cars~\cite{krause20133d}}&
        \multicolumn{3}{c}{FGVC-Aircraft~\cite{maji2013fine}}\\
        \cmidrule(lr{1em}){4-6} \cmidrule(lr{1em}){7-9} \cmidrule(lr{1em}){10-12}
        Global Relation&Local Relation&Feature Decoupling &All&Old&New&All&Old&New&All&Old&New\\
        \midrule
        \xmark&\xmark&\xmark &73.6&75.3&72.8 &58.5&75.6&50.3 &57.1&64.7&53.3\\
        \cmark&\xmark&\xmark &80.5&78.9&81.2 &61.9&81.0&52.6 &59.0&65.1&56.0\\
        \xmark&\cmark&\xmark &80.2&78.5&81.1 &64.3&82.3&55.7 &63.0&66.7&61.1\\
        \cmark&\cmark&\xmark &81.0&78.5&82.2 &65.8&82.8&57.6 &64.9&\textbf{68.4}&63.2\\
        \cmark&\cmark&\cmark &\textbf{82.8}&\textbf{80.3}&\textbf{84.1} &\textbf{66.5}&\textbf{83.4}&\textbf{58.3} &\textbf{66.6}&68.1&\textbf{65.8}\\

        \bottomrule
\end{tabular}
}
}
\label{tab:distillation}
\end{table}

\begin{table*}[t]
\centering
\caption{Experimental results using different backbone sizes with DINOv2~\cite{oquab2023dinov2} on SimGCD and SimGCD+CloSeR. Results are reported on SSB~\cite{vaze2022semantic}.}
\setlength{\tabcolsep}{1.3mm}{
\resizebox{\textwidth}{!}{
\begin{tabular}{ccc>{\columncolor{my_blue}}ccc>{\columncolor{my_blue}}ccc>{\columncolor{my_blue}}ccc>{\columncolor{my_blue}}ccc>{\columncolor{my_blue}}ccc>{\columncolor{my_blue}}ccc}
        \toprule
        &&&\multicolumn{3}{c}{CUB~\cite{wah2011caltech}}&
        \multicolumn{3}{c}{Stanford-Cars~\cite{krause20133d}}&
        \multicolumn{3}{c}{FGVC-Aircraft~\cite{maji2013fine}}&
        \multicolumn{3}{c}{CIFAR-10~\cite{krizhevsky2009learning}}&
        \multicolumn{3}{c}{CIFAR-100~\cite{krizhevsky2009learning}}&
        \multicolumn{3}{c}{ImageNet-100~\cite{deng2009imagenet}}\\
        \cmidrule(lr{1em}){4-6} \cmidrule(lr{1em}){7-9} \cmidrule(lr{1em}){10-12} \cmidrule(lr{1em}){13-15} \cmidrule(lr{1em}){16-18} \cmidrule(lr{1em}){19-21}
        Method&Teacher&Student  &All&Old&New&All&Old&New&All&Old&New&All&Old&New&All&Old&New&All&Old&New\\
        \midrule
        SimGCD&-&ViT-B &71.5&78.1&68.3 &71.5&81.9&66.6 &63.9&69.9&60.9 &98.7&96.7&99.7 &88.5&89.2&87.2 &89.9&95.5&87.1\\
        + CloSeR&ViT-B&ViT-B &78.1&79.5&77.3 &78.8&90.7&73.0 &71.8&68.7&73.3 &98.9&97.1&\textbf{99.8} &93.2&93.0 &\textbf{93.5} &92.5&\textbf{96.6}&90.4\\
        + CloSeR&ViT-L&ViT-B &\textbf{81.3}&\textbf{81.4}&\textbf{81.3} &\textbf{79.9}&\textbf{92.5}&\textbf{73.8}	&\textbf{75.0}&\textbf{77.7}&\textbf{73.6} &\textbf{99.4}&\textbf{98.9}&99.7 &\textbf{93.8}&\textbf{94.2}&93.0 &\textbf{93.4}&95.8&\textbf{92.2}\\
        \midrule
        SimGCD&-&ViT-L &71.0&73.7&69.6 &71.3&80.8&66.7 &70.6&70.1&70.8 &99.5&98.7&99.8 &91.8&91.8&91.8 &92.4&94.0&91.7\\
        + CloSeR&ViT-L&ViT-L &\textbf{81.4}&\textbf{83.2}&\textbf{80.5}	&\textbf{83.8}&\textbf{92.9}&\textbf{79.4} &\textbf{76.2}&\textbf{82.4}&\textbf{73.0} &\textbf{99.5}&\textbf{98.7}&\textbf{99.9} &\textbf{94.5}&\textbf{94.9}&\textbf{93.6}	 &\textbf{93.7}&\textbf{97.2}&\textbf{91.9}\\
        \bottomrule
\end{tabular}
}
}
\label{tab:model_size}
\end{table*}

\subsection{Diagnostic Study}
\noindent\textbf{Different Closed-Set Transfer Learning Methods.}
We compare three transfer strategies in the CST stage while keeping the rest of the CloSeR pipeline fixed: (1) full fine-tuning, (2) fine-tuning only the last transformer block (a standard practice in GCD training), and (3) our block-wise adapter adaptation.
For (1) and (2), we use an initial learning rate of $10^{-3}$ to reduce overfitting risk, whereas for (3), we use $10^{-1}$ given the much smaller set of trainable parameters.
As shown in Tab.~\ref{tab:adaptation}, full fine-tuning yields reasonable accuracy, but fine-tuning only the last block causes a pronounced performance drop, especially with DINO pretraining (\eg, Cars \textit{All} $60.9\!\rightarrow\!46.7$; Aircraft \textit{All} $54.8\!\rightarrow\!45.8$).
In contrast, block-wise adapters achieve the best overall results, consistently improving \textit{All} accuracy on DINO (CUB $68.1\!\rightarrow\!68.2$, Cars $60.9\!\rightarrow\!63.0$, Aircraft $54.8\!\rightarrow\!56.6$) while substantially strengthening \textit{New}-class discovery (\eg, Cars \textit{New} $52.9\!\rightarrow\!54.7$).
These results support two key observations.
First, it is beneficial to preserve the strong pretrained foundation weights rather than fully updating them on limited labelled data.
Second, effective transfer requires adapting not only high-level semantics but also \emph{shallow} features across the network hierarchy; this is particularly important for weaker pretraining (\eg, DINO), where tuning only the last block is insufficient to close the domain gap.

\noindent\textbf{Ablations on Knowledge Distillation Components.}
Tab.~\ref{tab:distillation} ablates the three components in URD with SelEx as baseline: {global} relation distillation (sample-to-prototype), {local} relation distillation (sample-to-sample), and {feature decoupling}.
Starting from the baseline without distillation, enabling either global or local relations yields large gains across all datasets, confirming the effectiveness of transferring relational structure from the closed-set teacher.
Local relations are particularly important on GCD datasets, improving \textit{All} accuracy from $58.5\!\rightarrow\!64.3$ on Cars and $57.1\!\rightarrow\!63.0$ on Aircraft, and outperforming using global relations alone on these datasets.
Importantly, global and local relations are complementary: combining them further improves performance (\eg, CUB \textit{All} $80.5/80.2\!\rightarrow\!81.0$, Cars $61.9/64.3\!\rightarrow\!65.8$, Aircraft $59.0/63.0\!\rightarrow\!64.9$).
Finally, feature decoupling brings an additional and consistent boost on top of global+local distillation.
Concretely, we decouple the features used by the two relation pathways: global relations are distilled in the closed-set prototype space using mapped student features, whereas local relations are distilled directly from the student backbone features.
This design avoids forcing a single representation to simultaneously satisfy prototype anchoring and neighborhood matching, reducing optimization interference and improving \textit{All} accuracy from $81.0\!\rightarrow\!82.8$ on CUB, $65.8\!\rightarrow\!66.5$ on Cars, and $64.9\!\rightarrow\!66.6$ on Aircraft.

\noindent\textbf{Scaling Up Teacher and Student Backbones.}
Tab.~\ref{tab:model_size} studies how performance changes with different teacher/student backbone sizes under DINOv2 pretraining.
A key finding is that the \emph{default} GCD training strategy in SimGCD does not reliably benefit from scaling up the backbone on fine-grained datasets: replacing a ViT-B student with a larger ViT-L model yields no improvement on CUB (\textit{All} $71.5\!\rightarrow\!71.0$) and Cars ($71.5\!\rightarrow\!71.3$).
This inconsistency suggests that directly optimizing the GCD objective can partially \emph{override} the strong pretrained priors of foundation models, preventing larger backbones from translating additional capacity into better clustering and recognition.
In contrast, CloSeR restores a more predictable scaling behavior.
With our closed-set teacher and URD supervision, increasing the student from ViT-B to ViT-L (with a ViT-L teacher) improves results across all three fine-grained datasets, especially on Cars (\textit{All} $79.9\!\rightarrow\!83.8$), while also improving on CUB ($81.3\!\rightarrow\!81.4$) and Aircraft ($75.0\!\rightarrow\!76.2$).
Moreover, CloSeR also benefits from a stronger teacher: using a ViT-L teacher improves over a ViT-B teacher with a fixed ViT-B student (\eg, CUB \textit{All} $78.1\!\rightarrow\!81.3$).
Overall, these results indicate that by constructing a domain-adapted closed-set teacher and distilling both global and local relations, CloSeR better preserves and exploits the foundation model prior, thereby \emph{releasing the power} of larger backbones for category discovery.

\begin{figure*}[t]
    \centering
    \includegraphics[width=\textwidth]{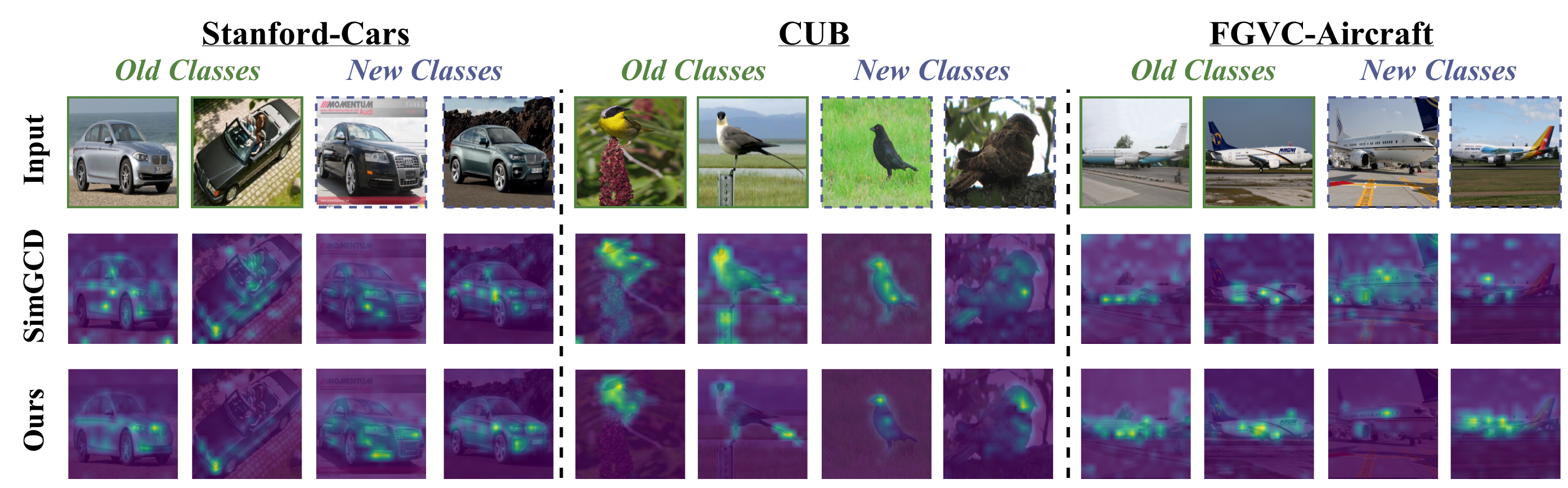}
    \caption{Comparison of attention maps from the backbone networks of SimGCD~\cite{wen2023parametric} and our SimGCD+CloSeR. The results include three datasets: Stanford-Cars~\cite{krause20133d}, CUB~\cite{wah2011caltech}, and FGVC-Aircraft~\cite{maji2013fine}, spanning both \emph{Old} and \emph{New} classes.}
    \label{fig:attn_vis}
\end{figure*}

\subsection{Qualitative Comparison}
Attention quality is crucial for fine-grained GCD~\cite{vaze2022generalized,wang2024sptnet,zhang2025less}: models must focus on category-discriminative object parts rather than task-irrelevant background, especially for novel classes where spurious cues can mislead clustering. 
To examine whether CloSeR improves visual focus, we visualize attention maps of SimGCD~\cite{wen2023parametric} and SimGCD+CloSeR in Fig.~\ref{fig:attn_vis}, derived from the last transformer block of the backbone across three fine-grained datasets.
Following~\cite{caron2021emerging}, we average attention over heads and bilinearly upsample the resulting heatmap to the input resolution, then normalize it to $[0,1]$ for overlay.
Qualitatively, our method focuses on semantically diagnostic parts and improves multi-part coverage (\eg, beaks/wing patterns in CUB, grilles/headlights in Stanford-Cars, body/wing markings in FGVC-Aircraft), while the baseline shows diffuse attention with background leakage, particularly on unseen categories.
Our method also improves multi-part coverage by highlighting complementary object parts, indicating stronger part-level reasoning and better generalization to novel categories, which supports improved accuracy in the GCD setting.

%% file: secs/5_conclusion.tex
\section{Conclusion}
\label{sec:conclusion}
We present CloSeR, a simple yet effective framework to improve generalized category discovery by leveraging foundation models through a \emph{closed-set} transfer learning stage and a unified \emph{relational} distillation stage.
First, we build a domain-adapted closed-set teacher via block-wise adapter tuning, which preserves the strong pretrained knowledge while adapting shallow-to-deep features efficiently.
Second, we introduce {Unified Relational Distillation (URD)} that transfers both prototype-anchored global relations and batch-level local neighborhood structure to the GCD student, with feature decoupling to reduce optimization interference.
Extensive experiments on fine-grained and generic benchmarks demonstrate consistent gains over strong baselines, and our ablations show that CloSeR provides a favorable accuracy--efficiency trade-off and better exploits model scaling, effectively unlocking the potential of larger backbones.